\documentclass[acmsmall,authorversion]{acmart}
\usepackage{comment}
\usepackage{bm}
\usepackage{booktabs}
\usepackage{siunitx}
\usepackage{subcaption}
\usepackage{algorithm}
\usepackage{algorithmic}
\usepackage{cleveref}
\renewcommand\footnotetextcopyrightpermission[1]{}
\setcopyright{none}
\acmConference{}{}{}
\acmYear{}
\acmDOI{}
\acmISBN{}
\title{Gradient Dynamics of Attention: How Cross-Entropy Sculpts Bayesian Manifolds}
\subtitle{Paper II of the Bayesian Attention Trilogy}
\author{Naman Agarwal}
\affiliation{
  \institution{Dream Sports}
  \city{New York}
  \state{NY}
  \country{USA}
}
\email{naman33k@gmail.com}
\authornote{Currently at Google DeepMind. Work performed while at Dream Sports.}
\author{Siddhartha R. Dalal}
\affiliation{
  \institution{Columbia University}
  \department{School of Professional Studies and Department of Statistics}
  \city{New York}
  \state{NY}
  \country{USA}
}
\email{sd2803@columbia.edu}
\author{Vishal Misra}
\affiliation{
  \institution{Columbia University}
  \department{Department of Computer Science}
  \city{New York}
  \state{NY}
  \country{USA}
}
\email{vishal.misra@columbia.edu}
\authorsaddresses{Authors' emails: naman33k@gmail.com, sd2803@columbia.edu, vishal.misra@columbia.edu. Authors are listed in alphabetical order.}
\begin{document}
\begin{abstract}
Our companion paper (Paper~I) establishes that neural sequence models can implement exact Bayesian inference, with success depending on whether the architecture realizes the required \emph{inference primitives}: belief accumulation, belief transport, and random-access binding. But \emph{how} does gradient descent learn to implement these primitives? We provide a systematic first-order analysis of how cross-entropy training reshapes attention scores and value vectors. Our core result is an \emph{advantage-based routing gradient} for attention scores,
\[
\frac{\partial L}{\partial s_{ij}}
= \alpha_{ij}\bigl(b_{ij}-\mathbb{E}_{\alpha_i}[b]\bigr),
\qquad
b_{ij} := u_i^\top v_j,
\]
coupled with a \emph{responsibility-weighted update} for values $v_j$,
\[
\Delta v_j = -\eta\sum_i \alpha_{ij} u_i,
\]
where $u_i$ is the upstream gradient at position $i$ and $\alpha_{ij}$ are attention weights. These equations induce a positive feedback loop in which routing and content specialize together: queries route more strongly to values that provide above-average loss reduction for their error signal, and those values are updated to oppose the error signals of the queries that use them.
We show that this coupled specialization behaves like a two-timescale EM procedure: attention weights implement an E-step (soft responsibilities), while values implement an M-step (responsibility-weighted prototype updates). This EM-like dynamic is what enables the inference primitives: belief accumulation emerges from responsibility-weighted value updates; belief transport emerges from content-dependent routing that tracks evolving states; and random-access binding emerges from the query-key matching that allows retrieval by content. Through controlled simulations, including a sticky Markov-chain task where we compare an EM-motivated learning schedule to standard SGD, we demonstrate that the same gradient dynamics that minimize cross-entropy also sculpt the low-dimensional Bayesian manifolds observed in Paper~I. We further propose an abstract framework for \emph{content-based value routing} that encompasses both attention and selective state-space models, conjecturing that advantage-based routing dynamics emerge in any architecture satisfying this framework---explaining why transformers and Mamba develop Bayesian geometry while LSTMs do not.
\end{abstract}
\maketitle
\section{Introduction}
Our companion paper (Paper~I) establishes that neural sequence models can implement exact Bayesian inference---filtering and hypothesis elimination---in controlled ``Bayesian wind tunnels.'' The key finding is that success depends on whether the architecture realizes the required \emph{inference primitives}: belief accumulation (integrating evidence), belief transport (propagating beliefs through dynamics), and random-access binding (retrieving hypotheses by content). Transformers realize all three primitives; Mamba realizes accumulation and transport; LSTMs realize only accumulation of static sufficient statistics; MLPs realize none. This taxonomy explains the empirical pattern: each architecture succeeds precisely on tasks demanding only the primitives it can implement.
But \emph{how} does gradient descent learn to implement these primitives? Why does cross-entropy training produce the geometric structures required for Bayesian inference---orthogonal key bases, progressive query alignment, low-dimensional value manifolds?
\vspace{4pt}
\noindent\textbf{This paper as Lemma 2 (Mechanism).}
Paper~I establishes \emph{existence}: transformers can implement exact Bayesian inference, and different architectures realize different primitives. This paper establishes \emph{mechanism}: when an architecture supports a given primitive, standard cross-entropy training induces gradient dynamics that reliably construct the corresponding probabilistic computation. Paper~III establishes \emph{scaling}: these mechanisms persist in production LLMs.
We analyze a single-head attention block trained with cross-entropy and derive all first-order gradients with respect to scores $s_{ij}$, queries $q_i$, keys $k_j$, and values $v_j$. We show that the resulting gradient dynamics implement an implicit EM-like algorithm, and that this EM structure is precisely what enables the inference primitives to emerge.
\subsection{Contributions}
Our main contributions are:
\begin{enumerate}
\item \textbf{Complete first-order analysis of attention gradients.}
We derive closed-form expressions for $\partial L/\partial s_{ij}$, $\partial L/\partial q_i$, $\partial L/\partial k_j$, and $\partial L/\partial v_j$ under cross-entropy loss, in a form that makes their geometric meaning transparent.
\item \textbf{Advantage-based routing law.}
We show that score gradients satisfy
\[
\frac{\partial L}{\partial s_{ij}}
= \alpha_{ij}\bigl(b_{ij}-\mathbb{E}_{\alpha_i}[b]\bigr),
\]
where $b_{ij} = u_i^\top v_j$ is a compatibility term. Since gradient descent subtracts the gradient, scores \emph{decrease} for positions whose compatibility higher than the current attention-weighted mean (positive gradient), and \emph{increase} for positions below average.
\item \textbf{Responsibility-weighted value updates and specialization.}
Values evolve according to
\[
\Delta v_j = -\eta \sum_i \alpha_{ij} u_i,
\]
a responsibility-weighted average of upstream gradients. This induces a positive feedback loop: queries route to values that help them; those values move toward their users, reinforcing routing and creating specialization.
\item \textbf{Two-timescale EM interpretation.}
We show that these dynamics implement an implicit EM-like algorithm: attention weights act as soft responsibilities (E-step), values as prototypes updated under those responsibilities (M-step), and queries/keys as parameters of the latent assignment model. Attention often stabilizes early, while values continue to refine---a frame--precision dissociation that matches our empirical observations in wind tunnels and large models.
\item \textbf{Toy experiments and EM vs. SGD comparison.}
In synthetic tasks, including a sticky Markov-chain sequence, we compare an EM-motivated learning rate schedule (with larger LR for values) to standard SGD. The EM-like schedule reaches low loss, high accuracy, and sharp predictive entropy significantly faster; SGD converges to similar solutions but with slower and more diffuse routing. PCA visualizations of value trajectories reveal emergent low-dimensional manifolds.
\end{enumerate}
Taken together with \citep{concurrent_bayes}, our results provide a unified story:
\begin{center}
\emph{Gradient descent} $\;\Rightarrow\;$ \emph{Bayesian manifolds} $\;\Rightarrow\;$ \emph{In-context inference.}
\end{center}
\noindent\textbf{Clarification on ``Bayesian inference.''}
Throughout this paper, ``Bayesian inference'' refers to the \emph{Bayesian posterior predictive over latent task variables}---not a posterior over network weights. Specifically, we address filtering and hypothesis elimination in tasks whose posteriors factorize sequentially, not general Bayesian model selection. We show that cross-entropy training sculpts geometry that implements these computations over in-context hypotheses.
\section{Setup and Notation}
We analyze a single attention head operating on a sequence of length $T$.
Indices $i,j,k$ run from $1$ to $T$ unless stated otherwise.
\subsection{Forward Pass}
At each position $j$ we have an input embedding $x_j \in \mathbb{R}^{d_x}$.
Linear projections produce queries, keys, and values:
\begin{align}
q_i &= W_Q x_i \in \mathbb{R}^{d_k}, \\
k_j &= W_K x_j \in \mathbb{R}^{d_k}, \\
v_j &= W_V x_j \in \mathbb{R}^{d_v}.
\end{align}
Attention scores, weights, and context vectors are:
\begin{align}
s_{ij} &= \frac{q_i^\top k_j}{\sqrt{d_k}}, \\
\alpha_{ij} &= \frac{\exp(s_{ij})}{\sum_r \exp(s_{ir})}, \\
g_i &= \sum_{j=1}^T \alpha_{ij} v_j \in \mathbb{R}^{d_v}.
\end{align}
The output of the head is passed through an output projection to logits:
\begin{align}
\ell_i &= W_O g_i + b \in \mathbb{R}^C, \\
p_i &= \mathrm{softmax}(\ell_i),
\end{align}
and we train with cross-entropy loss
\begin{equation}
L = -\sum_{i=1}^T \log p_{i,y_i},
\end{equation}
where $y_i$ is the target class at position $i$.
\subsection{Auxiliary Quantities}
For compactness we define:
\begin{align}
u_i &:= \frac{\partial L}{\partial g_i}
= W_O^\top (p_i - e_{y_i}) \in \mathbb{R}^{d_v},
\\
b_{ij} &:= u_i^\top v_j \in \mathbb{R}, \\
\mathbb{E}_{\alpha_i}[b]
&:= \sum_{j=1}^T \alpha_{ij} b_{ij}.
\end{align}
Here $u_i$ is the upstream gradient at position $i$, indicating how $g_i$ should move to reduce the loss. The scalar $b_{ij}$ measures compatibility between the error signal $u_i$ and value $v_j$, and $\mathbb{E}_{\alpha_i}[b]$ is the attention-weighted mean compatibility for query $i$.
\section{First-Order Gradient Derivations}
\label{sec:gradients}
We now derive all relevant gradients without skipping steps, focusing on forms that reveal their geometric meaning.
\subsection{Output Gradient}
For each $i$, the cross-entropy gradient with respect to logits is
\begin{equation}
\frac{\partial L}{\partial \ell_i} = p_i - e_{y_i}.
\end{equation}
This propagates back to the context $g_i$ as
\begin{equation}
u_i := \frac{\partial L}{\partial g_i}
= W_O^\top (p_i - e_{y_i}).
\end{equation}
Intuitively, $- u_i$ is the direction in value space along which moving $g_i$ would increase the logit of the correct token and decrease loss.
\subsection{Gradient with Respect to Values}
Since $g_i = \sum_j \alpha_{ij} v_j$, we have
\begin{align}
\frac{\partial L}{\partial v_j}
&= \sum_{i=1}^T \frac{\partial L}{\partial g_i}
\frac{\partial g_i}{\partial v_j}
= \sum_{i=1}^T u_i \alpha_{ij}.
\end{align}
Thus
\begin{equation}
\boxed{
\frac{\partial L}{\partial v_j}
= \sum_{i=1}^T \alpha_{ij}\, u_i.
}
\end{equation}
Geometrically, $v_j$ is pushed away from the attention-weighted average of upstream error vectors from all queries that use it.
\subsection{Gradient with Respect to Attention Weights}
Because $g_i$ depends linearly on $\alpha_{ij}$, we obtain
\begin{align}
\frac{\partial L}{\partial \alpha_{ij}}
&= \left(\frac{\partial L}{\partial g_i}\right)^\top
\frac{\partial g_i}{\partial \alpha_{ij}}
= u_i^\top v_j
= b_{ij}.
\end{align}
The scalar $b_{ij} = u_i^\top v_j$ measures the instantaneous compatibility between
the upstream gradient $u_i$ and value vector $v_j$. Since
$\partial L / \partial \alpha_{ij} = b_{ij}$, increasing $\alpha_{ij}$ increases
the loss when $b_{ij} > 0$ and decreases the loss when $b_{ij} < 0$.
Thus, $-b_{ij}$ represents the immediate marginal benefit (in loss reduction)
of allocating additional attention mass to position $j$ for query $i$.
\subsection{Gradient with Respect to Scores}
For fixed $i$, the softmax Jacobian is
\begin{equation}
\frac{\partial \alpha_{ij}}{\partial s_{ik}}
= \alpha_{ij}(\delta_{jk}-\alpha_{ik}).
\end{equation}
Applying the chain rule,
\begin{align}
\frac{\partial L}{\partial s_{ik}}
&= \sum_{j=1}^T \frac{\partial L}{\partial \alpha_{ij}}
\frac{\partial \alpha_{ij}}{\partial s_{ik}} \\
&= \sum_{j=1}^T b_{ij} \alpha_{ij}(\delta_{jk}-\alpha_{ik}) \\
&= \alpha_{ik} b_{ik}
- \alpha_{ik}\sum_{j=1}^T \alpha_{ij}b_{ij} \\
&= \alpha_{ik}\bigl(b_{ik}-\mathbb{E}_{\alpha_i}[b]\bigr).
\end{align}
Hence
\begin{equation}
\boxed{
\frac{\partial L}{\partial s_{ik}}
= \alpha_{ik}\Bigl(b_{ik}-\mathbb{E}_{\alpha_i}[b]\Bigr).
}
\label{eq:advantage-gradient}
\end{equation}
It is convenient to define an \emph{advantage} quantity with the sign chosen to
align with gradient descent:
\[
A_{ij} := -\bigl(b_{ij} - \mathbb{E}_{\alpha_i}[b]\bigr).
\]
Under this definition, $A_{ij} > 0$ indicates that increasing $s_{ij}$ (and hence
$\alpha_{ij}$) would reduce the loss. Gradient descent therefore increases
attention scores toward positions with positive advantage and decreases them
otherwise.
This is an \emph{advantage}-style gradient: since gradient descent subtracts the gradient, scores \emph{decrease} for positions whose compatibility $b_{ik}$ exceeds the attention-weighted average (positive gradient), and \emph{increase} for positions below average (negative gradient). Since high $b_{ik} = u_i^\top v_k$ indicates $v_k$ aligns with the error direction, this reallocates attention away from harmful values toward helpful ones.
\subsection{Gradients to $q_i$ and $k_j$}
We next propagate through the score definition $s_{ij} = q_i^\top k_j/\sqrt{d_k}$.
For fixed $i$,
\begin{align}
\frac{\partial L}{\partial q_i}
&= \sum_{k=1}^T \frac{\partial L}{\partial s_{ik}}
\frac{\partial s_{ik}}{\partial q_i}
= \sum_{k=1}^T \frac{\partial L}{\partial s_{ik}}
\frac{k_k}{\sqrt{d_k}} \\
&= \frac{1}{\sqrt{d_k}}
\sum_{k=1}^T
\alpha_{ik}\bigl(b_{ik}-\mathbb{E}_{\alpha_i}[b]\bigr) k_k.
\end{align}
Similarly, for fixed $j$,
\begin{align}
\frac{\partial L}{\partial k_j}
&= \sum_{i=1}^T \frac{\partial L}{\partial s_{ij}}
\frac{\partial s_{ij}}{\partial k_j}
= \sum_{i=1}^T \frac{\partial L}{\partial s_{ij}}
\frac{q_i}{\sqrt{d_k}} \\
&= \frac{1}{\sqrt{d_k}}
\sum_{i=1}^T
\alpha_{ij}\bigl(b_{ij}-\mathbb{E}_{\alpha_i}[b]\bigr) q_i.
\end{align}
Thus,
\begin{equation}
\boxed{
\begin{aligned}
\frac{\partial L}{\partial q_i}
&= \frac{1}{\sqrt{d_k}}
\sum_{k=1}^T
\alpha_{ik}\bigl(b_{ik}-\mathbb{E}_{\alpha_i}[b]\bigr) k_k, \\
\frac{\partial L}{\partial k_j}
&= \frac{1}{\sqrt{d_k}}
\sum_{i=1}^T
\alpha_{ij}\bigl(b_{ij}-\mathbb{E}_{\alpha_i}[b]\bigr) q_i.
\end{aligned}
}
\end{equation}
Finally, gradients propagate to $W_Q$ and $W_K$ as
\begin{align}
\nabla_{W_Q}L &= \sum_{i=1}^T
\left(\frac{\partial L}{\partial q_i}\right) x_i^\top, \\
\nabla_{W_K}L &= \sum_{j=1}^T
\left(\frac{\partial L}{\partial k_j}\right) x_j^\top.
\end{align}
\subsection{Gradient to $W_V$}
From $\partial L/\partial v_j = \sum_i \alpha_{ij} u_i$ and $v_j = W_V x_j$,
\begin{align}
\frac{\partial L}{\partial W_V}
&= \sum_{j=1}^T
\left(\frac{\partial L}{\partial v_j}\right) x_j^\top
= \sum_{j=1}^T
\Bigl(\sum_{i=1}^T \alpha_{ij} u_i\Bigr) x_j^\top.
\end{align}
Letting $U=[u_1,\dots,u_T]\in\mathbb{R}^{d_v\times T}$,
$A=(\alpha_{ij})\in\mathbb{R}^{T\times T}$,
and $X=[x_1,\dots,x_T]\in\mathbb{R}^{d_x\times T}$, we can write
\begin{equation}
\boxed{
\nabla_{W_V}L = UAX^\top.
}
\end{equation}
\subsection{First-Order Parameter Updates}
For learning rate $\eta>0$, gradient descent gives
\begin{align}
\Delta v_j &= -\eta \frac{\partial L}{\partial v_j}
= -\eta \sum_{i=1}^T \alpha_{ij} u_i,
\label{eq:delta-vj}
\\
\Delta s_{ik} &= -\eta \frac{\partial L}{\partial s_{ik}}
= -\eta \alpha_{ik}
\bigl(b_{ik}-\mathbb{E}_{\alpha_i}[b]\bigr),
\\
\Delta q_i &= -\eta \frac{\partial L}{\partial q_i}, \\
\Delta k_j &= -\eta \frac{\partial L}{\partial k_j}.
\end{align}
The remainder of the paper analyzes the coupled dynamics induced by these updates.
\section{Coupled Dynamics and Specialization}
\label{sec:dynamics}
We now unpack the implications of the gradient flows in \Cref{sec:gradients}, focusing on the interaction between routing (via scores and attention) and content (via values).
\subsection{Advantage-Based Attention Reallocation}
Equation~\eqref{eq:advantage-gradient} shows that for fixed query $i$, attention reallocates mass away from positions whose values are worse-than-average (i.e., $b_{ik} > \mathbb{E}_{\alpha_i}[b]$) for $u_i$ and toward those that are better-than-average. Recall the advantage $A_{ij} := -(b_{ij} - \mathbb{E}_{\alpha_i}[b])$ defined earlier: gradient descent increases attention scores toward positions with positive advantage, implementing an advantage-based routing rule.
\subsection{Value Updates as Responsibility-Weighted Prototypes}
Define the attention-weighted upstream signal for column $j$:
\begin{equation}
\bar{u}_j := \sum_{i=1}^T \alpha_{ij} u_i.
\end{equation}
From \eqref{eq:delta-vj},
\begin{equation}
\Delta v_j = -\eta \bar{u}_j,
\end{equation}
so $v_j$ moves in the direction that jointly benefits all queries that attend to it, weighted by their attention.
In particular, if a small set of queries uses $j$ heavily, $v_j$ becomes a prototype that serves that subset. This is the core specialization mechanism: values adapt to the error landscape created by their users.
\subsection{First-Order Effect on Contexts and Loss}
Since $g_i = \sum_k \alpha_{ik} v_k$, an infinitesimal change in $v_j$ induces
\begin{equation}
\Delta g_i = \alpha_{ij} \Delta v_j
= -\eta \alpha_{ij} \sum_{r=1}^T \alpha_{rj} u_r.
\end{equation}
Using $u_i = \partial L/\partial g_i$, the first-order change in loss from this perturbation is
\begin{align}
\Delta L_i &\approx u_i^\top \Delta g_i \\
&= -\eta \alpha_{ij} \sum_{r=1}^T \alpha_{rj} (u_i^\top u_r).
\label{eq:deltaL}
\end{align}
\paragraph{Interpretation.}
The factor $\alpha_{ij}$ says: the more query $i$ relies on value $j$, the more $v_j$'s update affects $g_i$. The column weights $\{\alpha_{rj}\}$ aggregate contributions from all queries that share $j$. The inner products $u_i^\top u_r$ measure whether their error directions are aligned (helpful) or opposed (conflicting).
If many $u_r$ align with $u_i$, then $v_j$ can move in a direction that helps them all. If some are anti-aligned, $v_j$ must compromise, and specialization may become harder.
\paragraph{Dominant-query approximation.}
If query $i$ is the primary user of key $j$ (i.e., $\alpha_{ij} \gg \alpha_{rj}$ for $r\neq i$), then
$\sum_r \alpha_{rj} u_r \approx \alpha_{ij}u_i$ and
\begin{align}
\Delta g_i &\approx -\eta \alpha_{ij}^2 u_i, \\
\Delta L_i &\approx -\eta \alpha_{ij}^2 \|u_i\|^2 \le 0.
\end{align}
Thus, under this approximation, the update decreases loss for query $i$ by an amount proportional to $\alpha_{ij}^2\|u_i\|^2$: the stronger the attention and the larger the error, the bigger the immediate improvement.
\subsection{Feedback Loop and Specialization}
Equations~\eqref{eq:advantage-gradient} and~\eqref{eq:delta-vj} together form a feedback loop:
\begin{enumerate}
\item If $v_j$ is particularly helpful to $i$ (large negative $b_{ij}-\mathbb{E}_{\alpha_i}[b]$, meaning $v_j$ points away from the error direction $u_i$), then $s_{ij}$ increases and $\alpha_{ij}$ grows.
\item Larger $\alpha_{ij}$ gives $u_i$ more weight in $\bar{u}_j$, so $v_j$ moves further away from $u_i$ (since $\Delta v_j = -\eta \bar{u}_j$).
\item As $v_j$ moves away from $u_i$, $b_{ij} = u_i^\top v_j$ decreases further, increasing $A_{ij}$ and reinforcing step~1.
\end{enumerate}
Repeated application yields specialized value vectors that serve distinct subsets of queries, while attention concentrates on these specialized prototypes.
\subsection{Geometric Illustration}
\begin{figure}[t]
\centering
\includegraphics[width=0.6\textwidth]{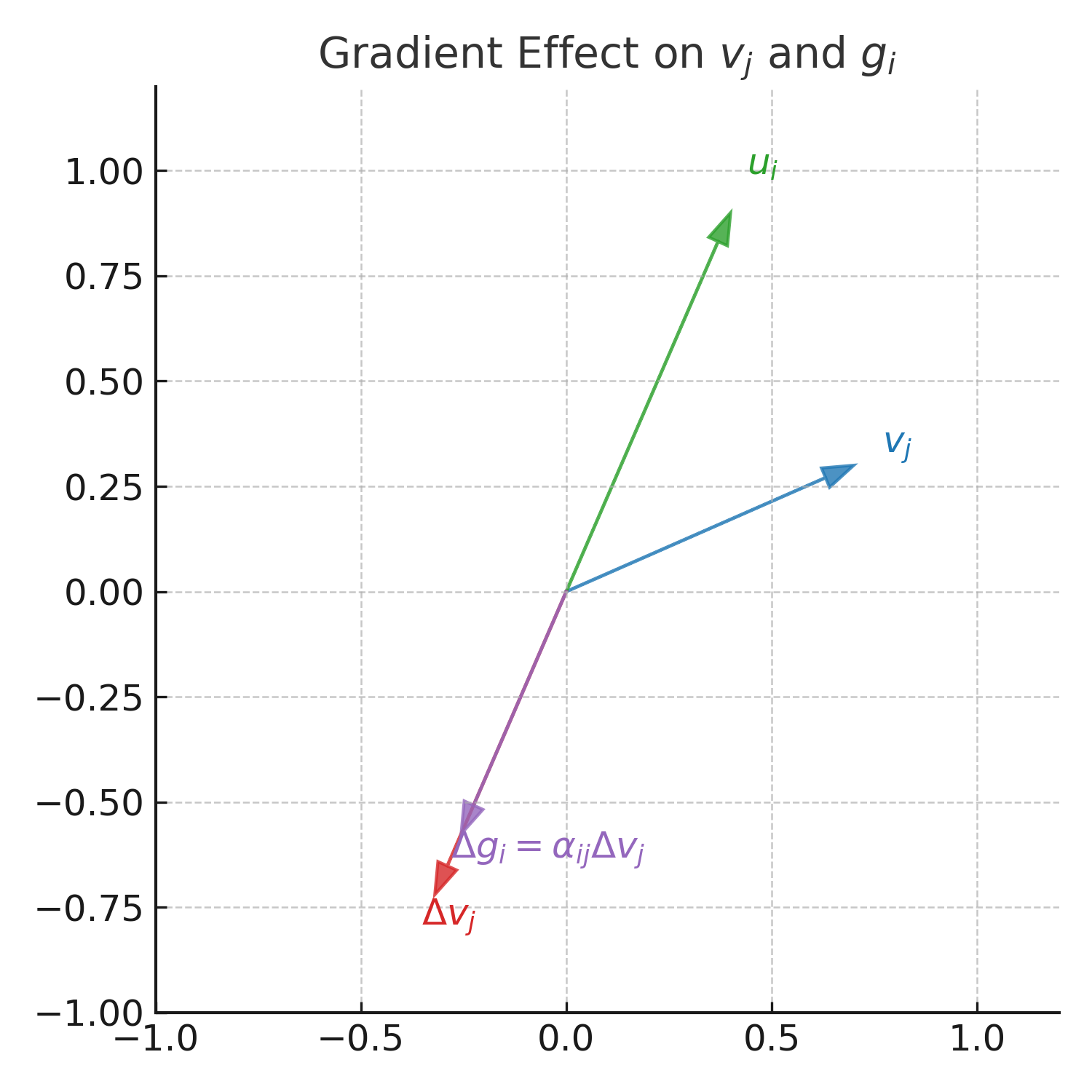}
\caption{Geometric illustration of coupled gradient dynamics. Value vector $v_j$ (blue) is updated away from the attention-weighted upstream signal $\bar{u}_j$ (green), inducing a change $\Delta g_i = \alpha_{ij}\Delta v_j$ (purple) that reduces loss and reinforces routing.}
\label{fig:gradient_effect}
\end{figure}

\Cref{fig:gradient_effect} schematically illustrates the coupled dynamics: $v_j$ is updated in the direction opposing upstream gradient $u_i$, inducing a change in $g_i$ along a direction that reduces loss; the improved alignment feeds back into the score gradients.
\section{EM-Like Two-Timescale Dynamics}
\label{sec:em}
The coupled dynamics derived above admit a useful analogy to
Expectation--Maximization (EM), not as a literal optimization of an explicit
latent-variable likelihood, but as a \emph{mechanistic correspondence} between
gradient flows and responsibility-weighted updates.
Attention weights behave like soft responsibilities over latent sources,
while value vectors act as prototypes updated under those responsibilities.
Unlike classical EM, the updates here are driven by upstream gradients rather
than observed data, and no standalone likelihood over values is optimized.
We emphasize that this is an interpretive framework: we do not derive a surrogate objective with guaranteed monotonic improvement. The value lies in the geometric intuition it provides for understanding specialization dynamics.
\subsection{Attention as Responsibilities}
For a fixed query $i$, the attention weights can be viewed as the posterior responsibilities of a latent variable $Z_i$ indicating which source position $j$ is active:
\begin{equation}
\alpha_{ij}
= p(Z_i = j \mid q_i, K; \theta)
= \frac{\exp(q_i^\top k_j/\sqrt{d_k})}
{\sum_{k} \exp(q_i^\top k_k/\sqrt{d_k})}.
\end{equation}
The score gradient
\[
\Delta s_{ij} \propto -\alpha_{ij}\bigl(b_{ij}-\mathbb{E}_{\alpha_i}[b]\bigr)
\]
decreases responsibilities for positions whose compatibility $b_{ij}$ exceeds the current mean, just as the E-step in EM increases responsibilities for mixture components that explain the data well.
\subsection{Values as Prototype Updates}
Given responsibilities $\alpha_{ij}$, the value update
\[
v_j^{\text{new}} = v_j^{\text{old}} - \eta \sum_i \alpha_{ij} u_i
\]
aggregates feedback from all assigned queries, weighted by their responsibilities. This is directly analogous to the M-step update for cluster means in Gaussian mixture models:
\[
\mu_j^{\text{new}}
\gets \frac{\sum_i \gamma_{ij} x_i}{\sum_i \gamma_{ij}},
\]
with $u_i$ playing the role of residuals that need to be explained.
A key difference from classical mixture-model EM is the role played by
$u_i$. In standard EM, centroids are updated using observed data points.
Here, $u_i$ represents the residual error signal backpropagated from the loss.
Values therefore move to better explain the \emph{error geometry} induced by the
current routing structure, rather than to maximize a likelihood over inputs.
The analogy is structural rather than variational.
\subsection{Approximate EM vs. SGD}
In classical EM, the E-step and M-step are separated: responsibilities are recomputed holding parameters fixed, then parameters are updated holding responsibilities fixed. In transformers trained with SGD, these steps are interleaved and noisy, but the first-order picture still resembles EM:
\begin{itemize}
\item \textbf{E-step (routing):} Score gradients adjust $q_i$ and $k_j$ so that attention patterns allocate responsibility to helpful positions. In practice, attention often stabilizes relatively early.
\item \textbf{M-step (values):} Value vectors continue to move under residual error signals, refining the content manifold even after attention appears frozen. This continues until calibration and likelihood converge.
\end{itemize}
Both updates occur simultaneously from each gradient computation. The ``two-timescale'' behavior emerges from relative gradient magnitudes and coupling structure, not from explicit alternation.
We can make this connection more explicit by considering a ``manual EM'' algorithm that alternates:
\begin{enumerate}
\item \emph{E-like step:} Forward pass to compute $\alpha_{ij}$ using current $q_i,k_j$.
\item \emph{M-like step on values:} Update $v_j$ with a closed-form gradient step using $\Delta v_j = -\eta\sum_i \alpha_{ij} u_i$ while freezing $q_i,k_j$.
\item \emph{Small SGD step on $W_Q,W_K,W_O$:} Update the nonlinear projection parameters with a small gradient step.
\end{enumerate}
In \Cref{sec:markov-sim}, we compare such an EM-like schedule to standard SGD on a sticky Markov-chain task and find that both converge to similar solutions, but the EM-like updates reach low loss and sharp, focused attention much faster.
\subsection{Analogy Table}
\begin{center}
\renewcommand{\arraystretch}{1.2}
\begin{tabular}{ll}
\toprule
EM Concept & Attention Equivalent \\
\midrule
Latent assignment $Z_i$ & Source index $j$ \\
Responsibilities $\gamma_{ij}$ & Attention weights $\alpha_{ij}$ \\
E-step & Forward pass + score update \eqref{eq:advantage-gradient} \\
Mixture means / centroids & Value vectors $v_j$ \\
M-step for means & $v_j \leftarrow v_j - \eta \sum_i \alpha_{ij} u_i$ \\
Convergence criterion & Cross-entropy loss, calibration, entropy \\
\bottomrule
\end{tabular}
\end{center}
\subsection{Bayesian vs. EM Perspective}
EM is an optimization procedure: it produces a point estimate $\theta^\ast$ maximizing (posterior) likelihood. A fully Bayesian treatment would instead integrate over $\theta$, but is intractable for transformers. Our analysis therefore lives at the EM/SGD level.
However, our companion work \citep{concurrent_bayes} shows that the point-estimate parameters learned in this way support \emph{Bayesian computation in representation space}: value manifolds, key frames, and query trajectories implement Bayesian belief updates in context. The present paper explains why cross-entropy and gradient descent naturally create these structures.
\section{From Gradient Flow to Bayesian Manifolds}
\label{sec:bayes-geometry}
We now connect the gradient dynamics derived above to the geometric structures observed in Bayesian wind tunnels and production models.
\subsection{Value Manifold Unfurling}
In wind-tunnel experiments \citep{concurrent_bayes}, which directly measure the geometric predictions made here:
\begin{itemize}
\item Early in training, attention entropy decreases and attention focuses on relevant hypotheses.
\item Later in training, attention patterns appear stable, but value representations unfurl along a smooth curve; PC1 explains 84--90\% of variance, strongly correlated with posterior entropy ($|r| > 0.9$).
\item Calibration error continues to drop even as attention maps remain visually unchanged.
\end{itemize}
Once score gradients have largely equalized compatibility ($b_{ij} \approx \mathbb{E}_{\alpha_i}[b]$), $\partial L/\partial s_{ij} \approx 0$ and attention freezes. But unless loss is exactly zero, $u_i$ remains non-zero and value updates
\[
\Delta v_j = -\eta \sum_i \alpha_{ij} u_i
\]
continue, gradually aligning $v_j$ along the principal directions of the residual error landscape. Under repeated updates, values come to lie on low-dimensional manifolds parameterized by downstream functionals such as posterior entropy.
\subsection{Hypothesis Frames and Key Orthogonality}
The gradients to $k_j$ show that keys are shaped by advantage signals:
\[
\frac{\partial L}{\partial k_j}
\propto \sum_i \alpha_{ij}
\bigl(b_{ij}-\mathbb{E}_{\alpha_i}[b]\bigr) q_i.
\]
If different subsets of queries consistently find different keys helpful, the corresponding gradient contributions push keys apart in $k$-space, encouraging approximate orthogonality of distinct hypothesis axes. Our wind-tunnel paper measures exactly this orthogonality: mean off-diagonal cosine similarity drops to 0.05--0.06 versus 0.08 for random vectors, a statistically significant reduction across multiple seeds ($p < 0.001$).
\subsection{Frame--Precision Dissociation}
The empirically observed \emph{frame--precision dissociation}---attention stably defining a hypothesis frame while calibration continues to improve---is now easy to interpret:
\begin{itemize}
\item The \emph{frame} is defined by $q_i,k_j$ and attention weights $\alpha_{ij}$. This stabilizes when advantage signals equalize and $\partial L/\partial s_{ij}\approx 0$.
\item \emph{Precision} lives in the arrangement of values $v_j$ along manifolds that support fine-grained likelihood modeling. This continues to refine as long as $u_i$ is non-zero.
\end{itemize}
Thus, a late-stage transformer has a fixed Bayesian frame (hypothesis axes and routing) but continues to sharpen its posterior geometry.
\subsection{Connection to the Dual-Entropy Framework}
\label{sec:dual-entropy}
The advantage-based routing gradient provides a mechanistic explanation for a dual-entropy measurement framework introduced in Paper~I~\citep{aggarwal2025bayesian1}.
Define \emph{context surprisal} $H_I(s) = -\log p_{\mathrm{train}}(s)$ as the distinctiveness of context $s$, and \emph{prediction entropy} $H_P(s)$ as the model's output uncertainty conditioned on $s$.
The ratio $\rho = H_P / H_I$ is a confidence-per-information coefficient: low $\rho$ from high $H_I$ indicates successful compositional inference; low $\rho$ from low $H_I$ indicates rote retrieval.
The advantage signal $b_{ij} - \mathbb{E}_{\alpha_i}[b]$ that drives routing is large precisely when $H_I$ is high and $H_P$ is low.
When context is distinctive (high $H_I$), the model encounters a pattern that strongly differentiates hypotheses, producing a large gradient on the attention scores.
When the model's prediction is already confident (low $H_P$), it successfully leverages that context to reduce loss---the advantage is realized.
The product of these two conditions yields strong, stable routing pressure: the E-step converges decisively, routing crystallizes, and the frame--precision dissociation proceeds.
When $H_I$ is low (common context), advantage signals are weak regardless of $H_P$---gradient pressure is mild and routing is slow.
When $H_I$ is high but $H_P$ remains high (the context is distinctive but the model cannot leverage it), the advantage signal is also weak: the model is not rewarded for attending to information it cannot use.
In both cases, $\rho$ stays high and routing fails to crystallize.
This connects the gradient-level mechanism (advantage-based routing) to the observable quantity ($\rho$): the EM-like dynamics drive $\rho$ down when and only when the routing target is stable across training episodes.
When the routing target is episode-dependent---as when token encodings change per episode or when input alignment varies---the advantage signal averages to noise, $\rho$ remains high throughout training, and no circuit compiles.
\section{Experiments}
\label{sec:experiments}
We now illustrate the theory with controlled simulations. All experiments use a single-head, single-layer attention block without residual connections or LayerNorm to keep the dynamics transparent.
\subsection{Toy Attention Simulation}
\begin{figure}[ht]
    \centering
    \begin{minipage}[b]{0.48\textwidth}
        \centering
        \includegraphics[width=\textwidth]{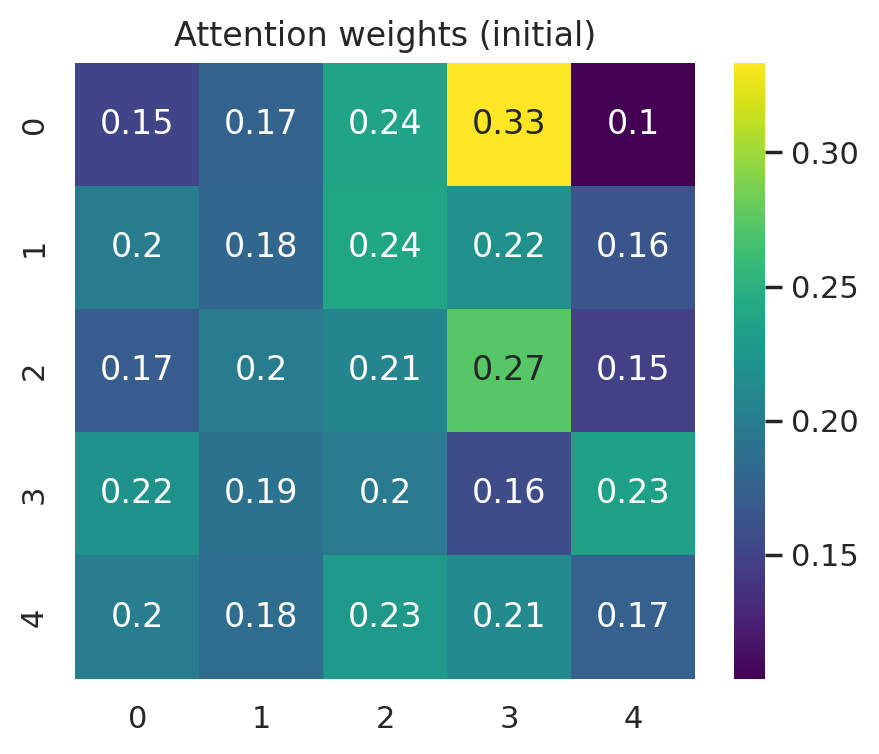}
        \caption{Initial attention heatmap (toy simulation)}
        \label{fig:heatmap_init}
    \end{minipage}
    \hfill 
    \begin{minipage}[b]{0.48\textwidth}
        \centering
        \includegraphics[width=\textwidth]{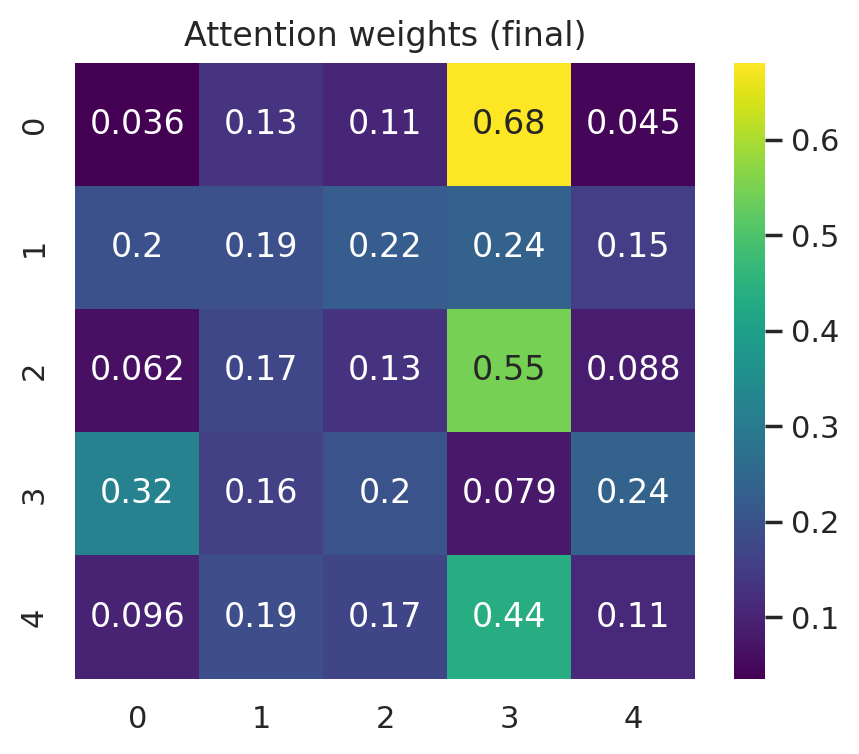}
        \caption{Final attention heatmap, 100 steps (toy simulation)} 
        \label{fig:heatmap_final}
    \end{minipage}
\end{figure}

\begin{figure}[ht]
    \centering
    \begin{minipage}[b]{0.48\textwidth}
        \centering
        \includegraphics[width=\textwidth]{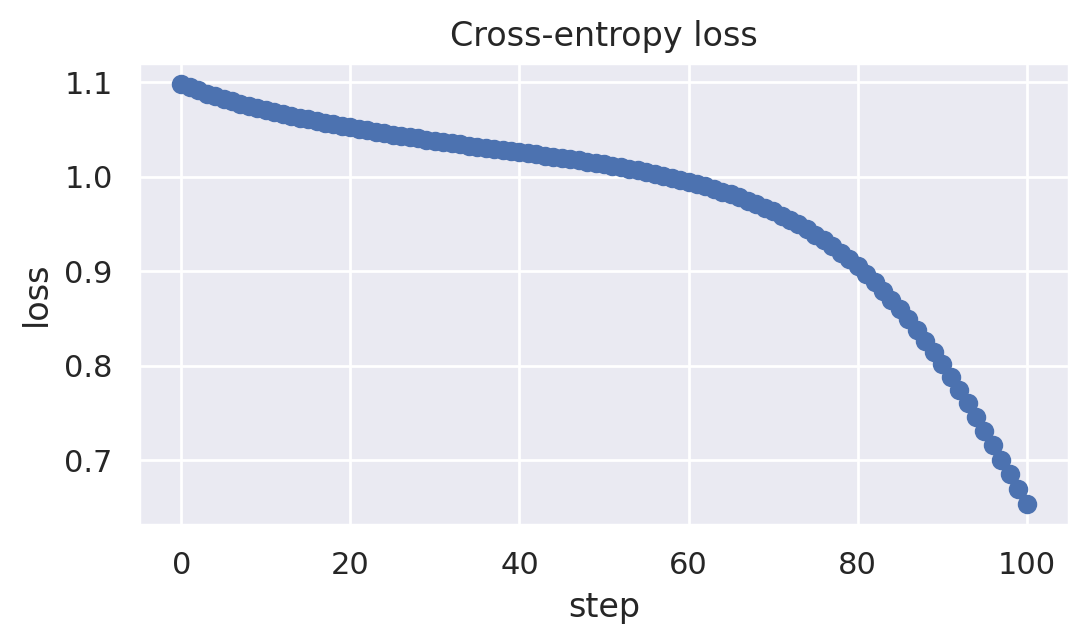}
        \caption{Loss, 100 EM steps (toy simulation)}
        \label{fig:loss_curve}
    \end{minipage}
    \hfill 
    \begin{minipage}[b]{0.3\textwidth}
        \centering
        \includegraphics[width=\textwidth]{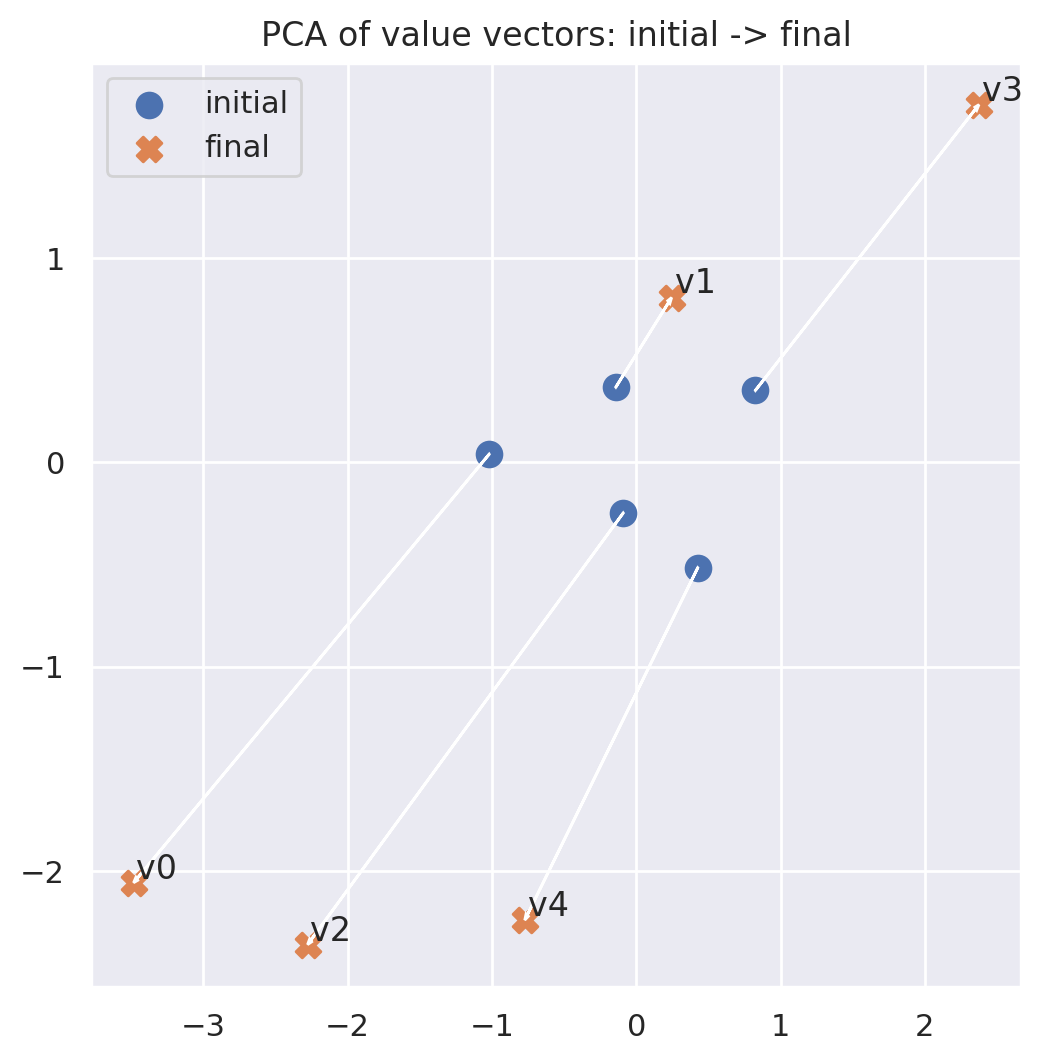}
        \caption{PCA projection of value vectors $v_j$ (toy simulation)}
        \label{fig:pca_dynamics}
    \end{minipage}
\end{figure}

We first consider a small toy setup with $T=5$, $d_x=3$, $d_k=2$, $d_v=2$, and $C=3$ classes.
\paragraph{Setup.}
\begin{itemize}
\item Projection matrices $W_Q,W_K,W_V,W_O$ are initialized with small Gaussian entries.
\item Inputs $x_j$ are drawn from $\mathcal{N}(0,I)$.
\item Targets $y_i$ are random in $\{0,1,2\}$.
\item We run manual update steps using the closed-form gradients from \Cref{sec:gradients}, with a fixed learning rate.
\end{itemize}
\paragraph{Observations.}
Over $\sim 100$ steps we observe:
\begin{enumerate}
\item Attention heatmaps sharpen: mass concentrates on a few positions per query (\Cref{fig:heatmap_init}, \Cref{fig:heatmap_final}).
\item Value vectors move coherently in a low-dimensional subspace; their trajectories in a PCA projection show emergent manifold structure (\Cref{fig:pca_dynamics}).
\item Cross-entropy loss decays smoothly (\Cref{fig:loss_curve}), with most gains occurring as specialization emerges.
\end{enumerate}
\subsection{Sticky Markov-Chain Simulation: EM vs. SGD}
\label{sec:markov-sim}
\begin{figure}[ht]
    \centering
    \begin{minipage}[b]{0.48\textwidth}
        \centering
        \includegraphics[width=\textwidth]{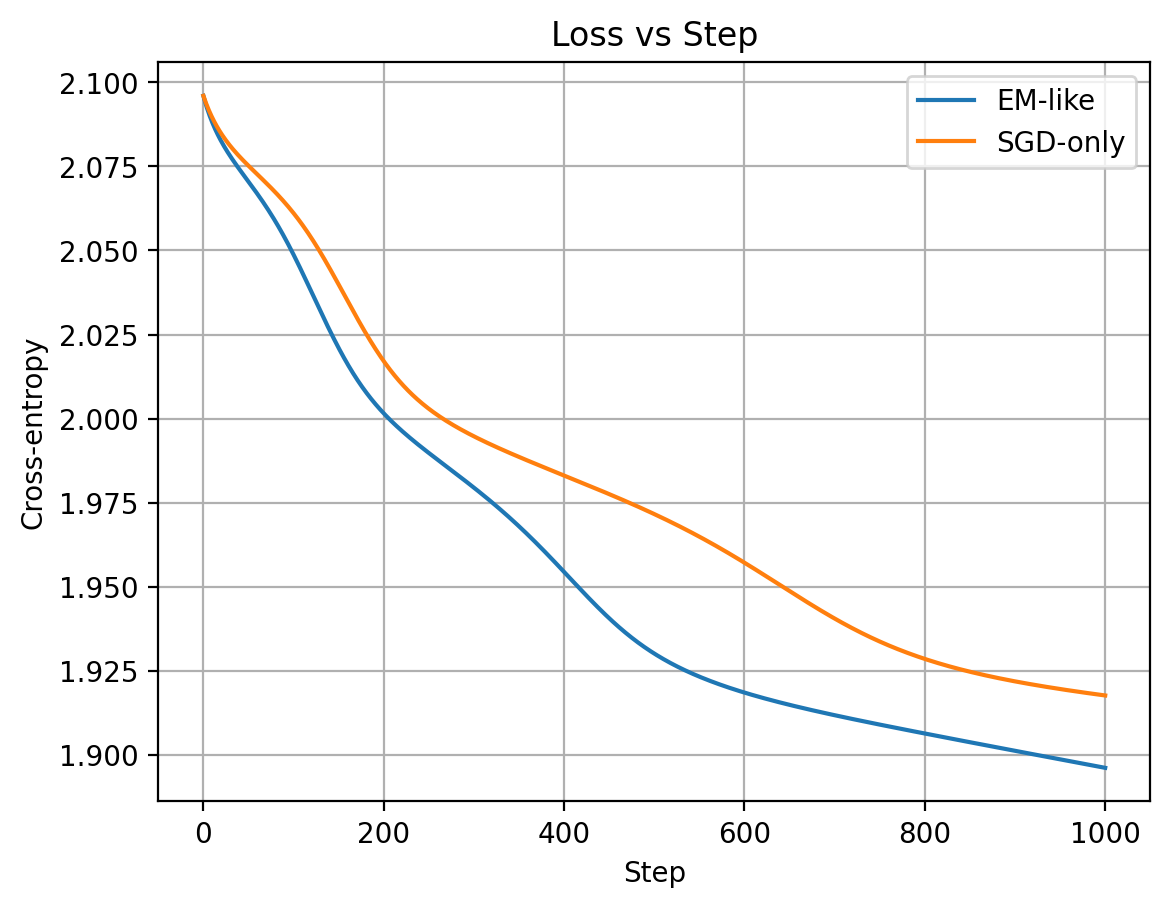}
        \caption{Sticky Markov Chain: Loss Dynamics}
        \label{fig:loss_sticky}
    \end{minipage}
    \hfill 
    \begin{minipage}[b]{0.48\textwidth}
        \centering
        \includegraphics[width=\textwidth]{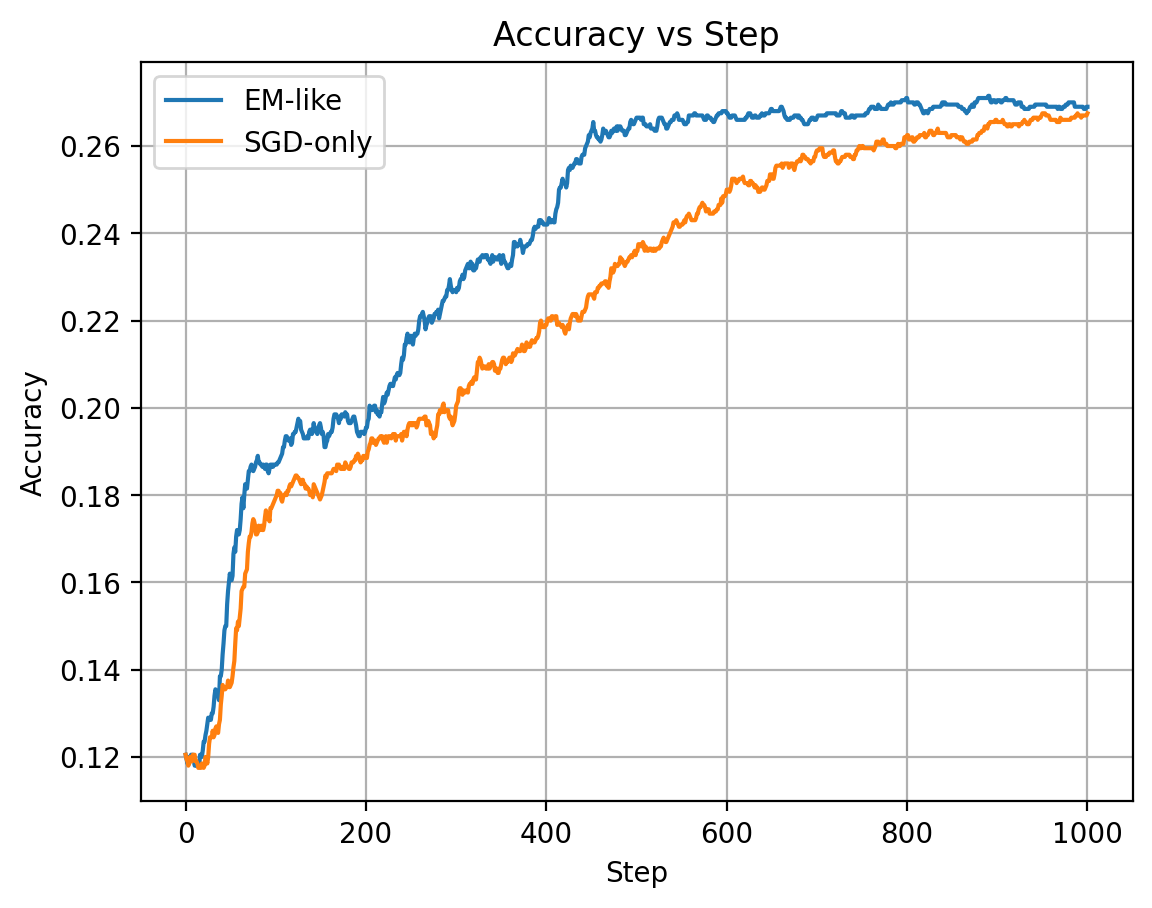}
        \caption{Sticky Markov Chain: Accuracy Dynamics}
        \label{fig:accuracy_sticky}
    \end{minipage}
\end{figure}

\begin{figure}[ht]
    \centering
    \begin{minipage}[b]{0.48\textwidth}
        \centering
        \includegraphics[width=\textwidth]{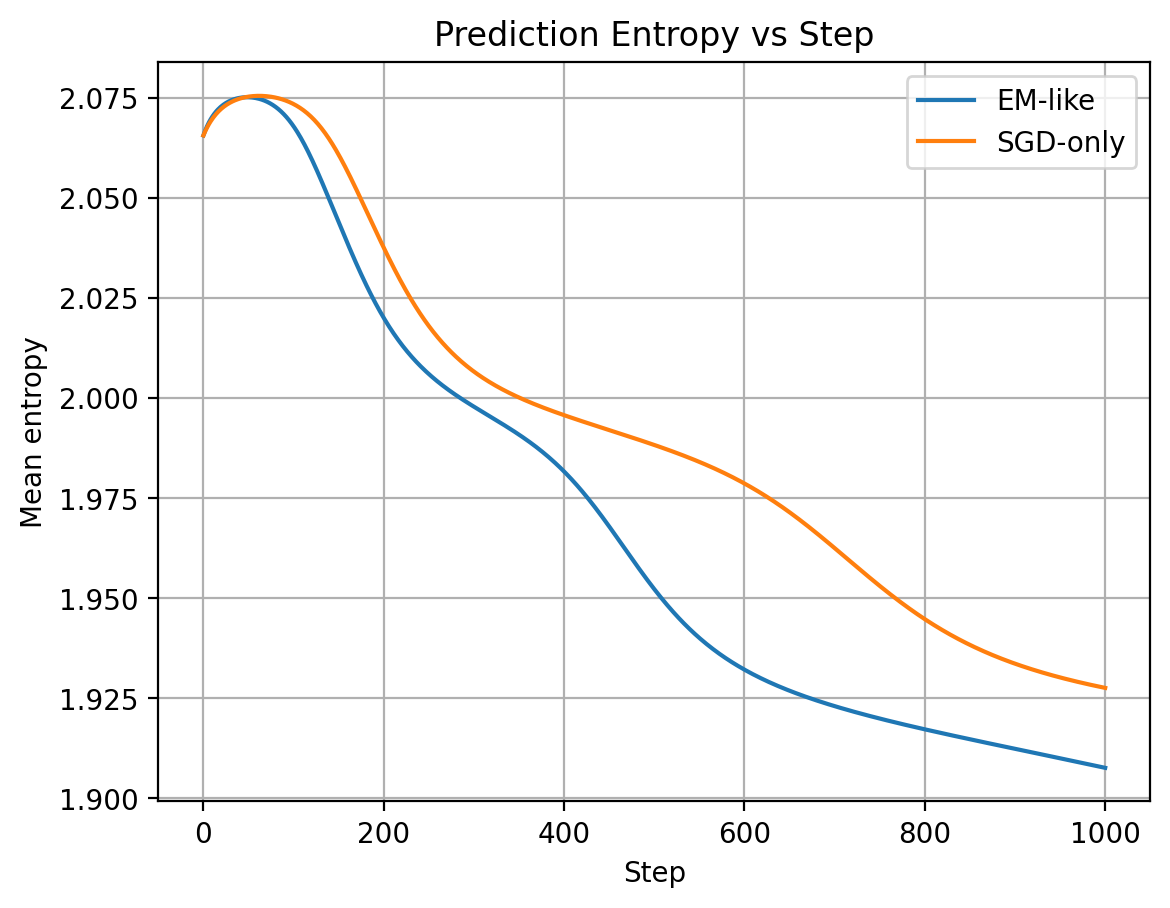}
        \caption{Sticky Markov Chain: Prediction Entropy}
        \label{fig:entropy_sticky}
    \end{minipage}
    \hfill 
    \begin{minipage}[b]{0.48\textwidth}
        \centering
        \includegraphics[width=\textwidth]{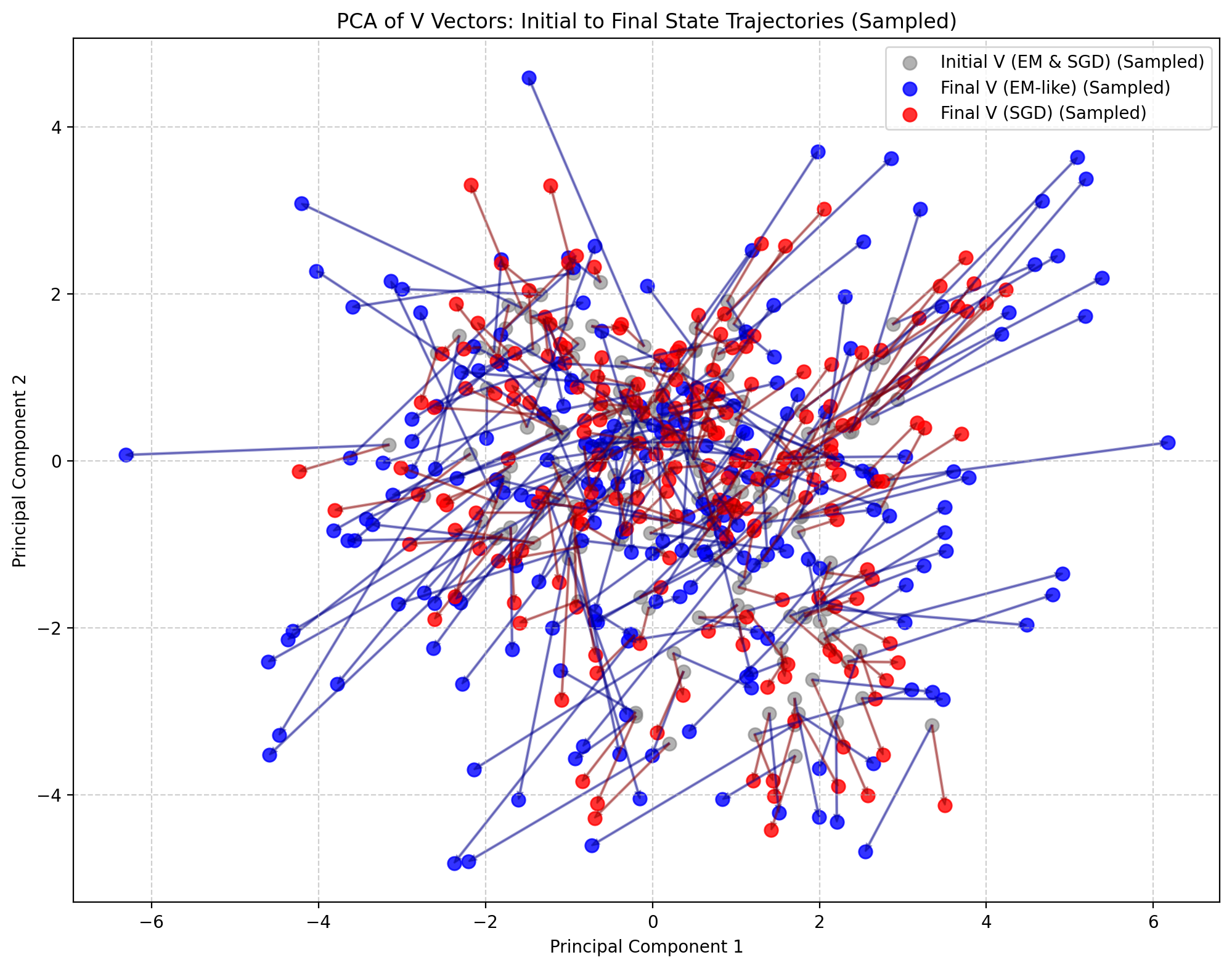}
        \caption{Sticky Markov Chain: 200 Sampled PCA of V } 
        \label{fig:pca_sticky}
    \end{minipage}
\end{figure}

We next study a more structured task where attention can exploit temporal persistence: a sticky Markov chain over symbols.
\paragraph{Task.}
We generate sequences of length $T=2000$ over an 8-symbol vocabulary $\{0,\dots,7\}$ from a first-order Markov chain with self-transition probability $P(y_{t+1}=y_t \mid y_t)=0.3$, otherwise transition to a different state with probability proportional to modulo distance of 7 with probability proportional to the distance from the previous state. With this circular distance, the nearby symbol gets higher probability than the distant one. Each symbol $y_t$ has an associated mean embedding in $\mathbb{R}^{20}$; the input at time $t$ is $x_t = \mu_{y_{t-1}} + \epsilon_t$ with $\epsilon_t \sim \mathcal{N}(0,I)$, so embeddings carry information about the previous symbol.
We train a single-head attention layer with $d_x=20$, $d_k=d_q=10$, $d_v=15$ for 1000 steps to predict $y_t$ from $(x_1,\dots,x_t)$.
\paragraph{Protocols.}
We compare two training schemes with causal masking and matched total parameter updates per step:
\begin{enumerate}
\item \textbf{Standard SGD.} Vanilla gradient descent on all parameters with learning rate $\eta = 0.01$.
\item \textbf{EM-like schedule.} Parameter-specific learning rates: $\eta_V = 0.1$ for value parameters, $\eta_{\text{routing}} = 0.01$ for $W_Q, W_K, W_O$. All updates are computed from a single forward-backward pass and applied simultaneously.
\end{enumerate}
The EM-like schedule uses a larger learning rate for values ($10\times$) to accelerate value specialization, compensating for the fact that value gradients are typically smaller in magnitude than routing gradients early in training. Both methods perform one gradient computation per step, with updates applied simultaneously (not alternating) using the respective learning rates.
\paragraph{Metrics.}
We track:
\begin{itemize}
\item final cross-entropy loss,
\item final accuracy,
\item predictive entropy,
\item $\mathrm{KL}(\text{EM} \parallel \text{SGD})$ of the predictive distributions.
\end{itemize}
\paragraph{Results.}
\Cref{tab:metrics} summarizes results across 5 random seeds after 1000 training steps.
\begin{table}[t]
    \centering
    \caption{Sticky Markov-chain task: EM-like vs.\ standard SGD (mean $\pm$ std over 5 seeds). EM achieves significantly lower loss (paired $t$-test, $p=0.003$), higher accuracy, and sharper predictive distributions. Critically, EM reaches SGD's final loss level in only $430 \pm 143$ steps---a $2.3\times$ convergence speedup. The theoretical Bayesian minimum entropy is computed from the transition matrix; actual minimum is slightly higher due to input noise.}
    \label{tab:metrics}
    \begin{tabular}{lcc}
        \toprule
        \textbf{Metric} & \textbf{EM-like} & \textbf{SGD} \\
        \midrule
        Final loss & $1.970 \pm 0.034$ & $2.058 \pm 0.007$ \\
        Final entropy & $1.998 \pm 0.035$ & $2.077 \pm 0.001$ \\
        Final accuracy & $0.244 \pm 0.016$ & $0.200 \pm 0.009$ \\
        \midrule
        Steps to SGD level & $430 \pm 143$ & 1000 \\
        Theoretical Min Entropy & \multicolumn{2}{c}{1.829} \\
        \bottomrule
    \end{tabular}
\end{table}
\Cref{fig:loss_sticky}--\Cref{fig:entropy_sticky} plot loss, accuracy, and entropy as a function of training step. EM-like training converges significantly faster than SGD: it reaches SGD's final loss in under half the steps ($2.3\times$ speedup), with loss and entropy dropping sharply while accuracy rises quickly. This acceleration occurs because the EM-motivated learning rate schedule (with larger LR for values) allows values to specialize more rapidly, whereas uniform SGD must balance routing and content updates at the same rate. Even with this single-layer, single-head architecture, both methods approach the theoretical Bayesian minimum (1.829 bits), with EM converging closer (1.998 vs 2.077 bits entropy). The residual gap of $\sim$0.17 bits from the theoretical minimum likely reflects the limited capacity of a single-head architecture to fully capture the transition structure; concurrent work \citep{concurrent_bayes} shows that deeper architectures close this gap to $<0.01$ bits on similar tasks.
To compare micro behavior of EM and SGD, we computed the movement of $v$ vectors for EM and SGD. Since $T=2000$, we will have 2000 $v_j$ for each method. For clearer visualization we show a sample of 200 $v$'s. The PCA projection plot of these 200 $v_j$ vector trajectories is quite insightful. It shows (\Cref{fig:pca_sticky}) that EM induces longer, more coherent trajectories from initial to final positions than SGD, consistent with stronger, more focused specialization. Red arrows (SGD) are shorter and more scattered; blue arrows (EM) are longer and more aligned, indicating that the EM-like updates exploit the analytic form of the value gradient more effectively.
\paragraph{Takeaways.}
Both EM-like and SGD training ultimately converge toward similar qualitative solutions: specialized values and focused attention. However, the EM-like schedule reaches this state with fewer steps and sharper specialization. This matches the two-timescale story: responsibilities (attention) can be treated as approximately converged, and closed-form value updates can exploit this stability to accelerate manifold formation.
\section{Practical Implications and Diagnostics}
\label{sec:practical}
The gradient analysis suggests useful diagnostics and design principles for training and interpreting transformer attention.
\subsection{Diagnostics}
\begin{itemize}
\item \textbf{Compatibility matrix $B=(b_{ij})$.} Monitoring $b_{ij} = u_i^\top v_j$ reveals which values are most helpful to which queries.
\item \textbf{Advantage matrix $A^{\text{adv}}=(b_{ij}-\mathbb{E}_{\alpha_i}[b])$.} The sign and magnitude predict where attention will strengthen or weaken.
\item \textbf{Column usage $\sum_i \alpha_{ij}$.} Low column sums identify underused or dead values.
\item \textbf{Value norms $\|v_j\|$.} Norm trajectories can flag potential instability (exploding or vanishing norms).
\end{itemize}
\subsection{Regularization and Stability}
\begin{itemize}
\item \textbf{LayerNorm on values} can stabilize norms while leaving directional dynamics intact.
\item \textbf{Attention dropout} disrupts the feedback loop, limiting over-specialization and encouraging more evenly used values.
\item \textbf{Learning rate choices} modulate the timescale separation between routing and content; smaller learning rates make the first-order picture more accurate.
\end{itemize}
\subsection{Architectural Choices}
\begin{itemize}
\item \textbf{Multi-head attention} allows multiple specialized routing manifolds, reducing competition within a single head.
\item \textbf{Depth} naturally supports the binding--elimination--refinement hierarchy observed in our wind-tunnel experiments and large models.
\item \textbf{Residual connections} help maintain useful intermediate representations even as individual heads specialize strongly.
\end{itemize}
\section{Toward a General Theory of Content-Based Routing}
\label{sec:general-routing}
The gradient dynamics derived in \Cref{sec:dynamics,sec:em} are specific to softmax attention.
However, companion work shows that selective state-space models (Mamba) achieve comparable
Bayesian inference through a different routing mechanism---input-dependent gating rather than
query-key matching. This raises a natural question: is there a unifying principle that explains
why both architectures develop Bayesian geometry?
We conjecture that the essential structure is \emph{content-based value routing}, and that
any mechanism satisfying this property will exhibit similar gradient dynamics under cross-entropy training.
\subsection{Content-Based Routing: An Abstract Definition}
\begin{definition}[Content-Based Value Routing]
A \emph{content-based value routing mechanism} is a parameterized function that, given a sequence
of representations $\{x_1, \ldots, x_T\}$, computes:
\begin{enumerate}
    \item \textbf{Routing weights} $w_{ij}(\theta; x_{1:T}) \in [0,1]$ that depend on the \emph{content}
    of positions $i$ and $j$ (not merely their indices), with $\sum_j w_{ij} = 1$.
    \item \textbf{Value representations} $v_j(\phi; x_j)$ computed from input content.
    \item \textbf{Output} $o_i = \sum_j w_{ij} v_j$, a routing-weighted combination of values.
\end{enumerate}
\end{definition}
The key requirement is that routing weights depend on \emph{what} is at each position, not just
\emph{where} it is. This excludes:
\begin{itemize}
    \item Fixed positional patterns (e.g., ``always attend to position $t-1$'')
    \item Learned but content-independent weights (e.g., sinusoidal position encodings alone)
    \item Architectures where gating is independent of cross-position relationships (e.g., standard LSTM gates)
\end{itemize}
\subsection{Instantiations}
\paragraph{Softmax Attention.}
The standard transformer attention mechanism is a content-based routing mechanism with:
\begin{align}
    w_{ij} &= \frac{\exp(q_i^\top k_j / \sqrt{d})}{\sum_{j'} \exp(q_i^\top k_{j'} / \sqrt{d})} \\
    v_j &= W_V x_j \\
    o_i &= \sum_j w_{ij} v_j
\end{align}
Routing weights depend on content through the query-key inner product $q_i^\top k_j$.
\paragraph{Selective State-Space Models (Mamba).}
Mamba implements content-based routing through input-dependent state-space parameters:
\begin{align}
    \Delta_t, B_t, C_t &= \text{proj}(x_t) \quad \text{(input-dependent)} \\
    h_t &= \exp(A \cdot \Delta_t) h_{t-1} + \Delta_t \cdot B_t \cdot x_t \\
    o_t &= C_t \cdot h_t
\end{align}
The matrices $\Delta_t$ (discretization step), $B_t$ (input projection), and $C_t$ (output projection)
are computed from the input $x_t$, making the routing content-dependent. The effective routing weight
from position $j$ to position $i > j$ is:
\[
w_{ij}^{\text{SSM}} = C_i \cdot \left(\prod_{k=j+1}^{i} \exp(A \cdot \Delta_k)\right) \cdot \Delta_j \cdot B_j
\]
This depends on the content at all intermediate positions, satisfying the content-based criterion.
\paragraph{LSTM (Counterexample).}
Standard LSTMs have gates that depend only on the current input and previous hidden state:
\begin{align}
    f_t &= \sigma(W_f [h_{t-1}, x_t]) \\
    i_t &= \sigma(W_i [h_{t-1}, x_t])
\end{align}
Crucially, the forget gate $f_t$ and input gate $i_t$ do not depend on the \emph{relationship}
between the current position's content and other positions' content. The LSTM cannot selectively
route based on content similarity across positions---it applies the same gating logic regardless
of what information is available elsewhere in the sequence. This explains why LSTMs can accumulate \emph{static} sufficient statistics (bijection elimination admits a fixed-dimensional statistic) but fail when the statistic must evolve under dynamics (HMM transport) or be indexed by content (binding).
\subsection{Conjectured General Gradient Dynamics}
We conjecture that any content-based routing mechanism, when trained with cross-entropy loss,
exhibits the following gradient structure:
\begin{conjecture}[Advantage-Based Routing]
\label{conj:advantage}
Let $w_{ij}(\theta)$ be content-based routing weights parameterized by $\theta$. Under cross-entropy
loss $L$, the gradient with respect to routing parameters satisfies:
\[
\frac{\partial L}{\partial \theta} \propto \sum_{i,j} w_{ij} \cdot \underbrace{(b_{ij} - \mathbb{E}_{w_i}[b])}_{\text{advantage}} \cdot \frac{\partial w_{ij}}{\partial \theta}
\]
where $b_{ij}$ measures the compatibility between the upstream gradient at $i$ and the value at $j$.
\end{conjecture}
\begin{conjecture}[Responsibility-Weighted Value Updates]
\label{conj:responsibility}
Value parameters $\phi$ evolve according to:
\[
\frac{\partial L}{\partial \phi} \propto \sum_{i,j} w_{ij} \cdot u_i \cdot \frac{\partial v_j}{\partial \phi}
\]
where $u_i$ is the upstream gradient. Values are pulled toward the queries/outputs that route to them,
weighted by routing strength.
\end{conjecture}
For softmax attention, we proved these conjectures exactly in \Cref{sec:dynamics}. For selective SSMs,
the sequential structure complicates the analysis (gradients must backpropagate through the recurrence),
but preliminary analysis suggests the same qualitative structure emerges.
\subsection{Implications for Bayesian Geometry}
If Conjectures~\ref{conj:advantage} and~\ref{conj:responsibility} hold generally, they explain why
both attention and selective SSMs develop Bayesian geometry:
\begin{enumerate}
    \item \textbf{Routing specialization}: Advantage-based updates cause routing to specialize---weights
    increase for value positions that are ``above average'' at reducing the loss for each query.
    \item \textbf{Value clustering}: Responsibility-weighted updates pull values toward the queries
    that use them, creating clusters in value space.
    \item \textbf{Manifold formation}: The combination produces low-dimensional value manifolds
    where position encodes task-relevant information (e.g., posterior entropy).
    \item \textbf{EM-like dynamics}: The coupled specialization resembles EM: routing weights act
    as soft responsibilities (E-step), values act as prototypes (M-step).
\end{enumerate}
This explains the empirical finding from Paper~1: transformers and Mamba both achieve Bayesian
inference (0.049 and 0.031 bits MAE on HMM tracking), while LSTMs fail (0.416 bits). The LSTM
lacks content-based routing and therefore cannot develop the coupled specialization that produces
Bayesian geometry.
\subsection{Future Work: Formal Analysis of Selective SSMs}
A complete theory requires deriving the gradient dynamics for selective SSMs explicitly. The key
challenges are:
\begin{enumerate}
    \item \textbf{Sequential dependencies}: Unlike attention, SSM gradients must backpropagate
    through the recurrence $h_t = f(h_{t-1}, x_t, \Delta_t, B_t)$.
    \item \textbf{Coupled parameters}: The input-dependent $\Delta$, $B$, $C$ are computed through
    shared projections, creating coupled dynamics.
    \item \textbf{Continuous vs.\ discrete}: SSMs operate in continuous state-space; the EM analogy
    may require a continuous relaxation (e.g., variational inference interpretation).
\end{enumerate}
We leave formal derivation for future work, but note that the empirical success of Mamba on
Bayesian tasks strongly suggests that similar dynamics emerge. The abstract framework proposed
here provides a roadmap for unifying the theory of content-based routing across architectures.
\section{Related Work}
\subsection{Bayesian Interpretations of Transformers}
Several works argue that transformers implement approximate Bayesian inference, either behaviorally or via probing \citep[e.g.][]{xie2022explanation,vonoswald2023transformers}. Our companion paper \citep{concurrent_bayes} demonstrates exact Bayesian behavior in small wind tunnels, showing that architectures with content-based value routing (transformers and Mamba) succeed while those without it (LSTMs, MLPs) fail. A scaling paper shows similar geometric signatures in production LLMs. The present work explains \emph{how} gradient dynamics in attention produce these geometries---deriving analogous dynamics for selective state-space models remains future work. 
\subsection{Mechanistic Interpretability}
Mechanistic interpretability studies identify specific heads and circuits performing copy, induction, and other algorithmic tasks \citep{elhage2021mathematical,olsson2022context}. Our framework complements this line by explaining how specialization arises from the interaction between routing and content, rather than treating specialized heads as primitive.
\subsection{Optimization and Implicit Bias}
The implicit bias of gradient descent in linear and deep networks has been extensively studied \citep{soudry2018implicit,ba2022high}. We extend these ideas to attention: gradient descent implicitly favors representations where routing aligns with error geometry and values lie on low-dimensional manifolds that support Bayesian updating.
\paragraph{Attention Training Dynamics.}
Recent work has analyzed attention optimization from complementary perspectives. Max-margin analyses show directional convergence of attention parameters toward SVM-like solutions. Two-stage dynamics have been identified where early training exhibits parameter condensation followed by key/query activation. Our advantage-based routing perspective complements these findings: the baseline-subtracted compatibility gradient we derive is the mechanism by which attention reallocates mass, and our two-timescale observation (fast routing, slow values) aligns with reports of stage-wise dynamics. The key difference is our focus on the EM-like interpretation and the connection to Bayesian manifold formation.
The responsibility-weighted value updates derived here are reminiscent of
neural EM and slot-attention models, where soft assignments drive prototype
updates. The key distinction is that, in transformers, responsibilities are
computed via content-addressable attention and prototype updates are driven by
backpropagated error signals rather than reconstruction likelihoods. Our focus
is not on proposing a new EM-style architecture, but on showing how standard
cross-entropy training in attention layers induces EM-like specialization
dynamics as a consequence of gradient flow.
\section{Limitations and Future Directions}
Our analysis is deliberately minimal and controlled.
\paragraph{First-order approximation.}
We work in a first-order regime, assuming small learning rates and ignoring higher-order and stochastic effects (e.g., momentum, Adam, mini-batch noise). Extending the analysis to realistic optimizers is an important next step.
\paragraph{Single-head, single-layer focus.}
We analyze a single head in isolation, without residual pathways or LayerNorm. Multi-head, multi-layer dynamics---including inter-head coordination and hierarchical specialization---remain open.
\paragraph{Finite vs.\ infinite width.}
We do not explicitly connect our analysis to the neural tangent kernel or infinite-width limits. Bridging these regimes may help clarify when transformers operate in a feature-learning versus lazy-training mode.
\paragraph{Large-scale empirical validation.}
Our toy simulations are intentionally small. Applying the diagnostics in \Cref{sec:practical} to full-scale LLM training runs, tracking advantage matrices and manifold formation over time, is a promising direction.
\paragraph{Attention-specific analysis.}
Our gradient derivations are specific to softmax attention. Companion work shows that selective state-space models (Mamba) also implement Bayesian inference via content-based routing, but through different mechanics---input-dependent gating rather than query-key matching. Deriving analogous gradient dynamics for selective SSMs, and identifying whether a similar EM-like structure emerges, is an important direction for unifying the theory of content-based routing across architectures.
\section{Conclusion}
Paper~I establishes \emph{what} architectures can implement Bayesian inference, taxonomized by three inference primitives: belief accumulation, belief transport, and random-access binding. This paper explains \emph{how} gradient descent learns to implement these primitives.
Our key findings are:
\begin{enumerate}
\item \textbf{Advantage-based routing.} Score gradients $\partial L/\partial s_{ij}
= \alpha_{ij}(b_{ij}-\mathbb{E}_{\alpha_i}[b])$ implement an advantage rule that reallocates attention toward positive-advantage values (those with below-average compatibility scores $b_{ij}$. This dynamic enables \emph{random-access binding}: the query-key mechanism learns to retrieve by content.
\item \textbf{Responsibility-weighted value updates.} Values evolve via $\Delta v_j = -\eta \sum_i \alpha_{ij} u_i$, becoming prototypes for the queries that attend to them. This enables \emph{belief accumulation}: values aggregate evidence from positions that route to them.
\item \textbf{Coupled specialization.} Routing and content form a positive feedback loop that produces specialized values and focused attention. This enables \emph{belief transport}: content-based routing can track how beliefs evolve through stochastic dynamics.
\item \textbf{Two-timescale EM behavior.} Attention often stabilizes early (E-step), while values refine more slowly (M-step), explaining the observed frame--precision dissociation.
\item \textbf{Manifold formation.} The same gradient dynamics that reduce cross-entropy loss also sculpt the low-dimensional manifolds that implement Bayesian posteriors in representation space.
\end{enumerate}
The EM interpretation explains \emph{why} the inference primitives emerge: the E-step (attention) learns what information to route where; the M-step (values) learns what beliefs to store. Architectures without content-based routing (LSTMs, MLPs) cannot implement the E-step and therefore cannot learn the coupled specialization that produces Bayesian geometry. LSTMs can accumulate static sufficient statistics through their recurrent state, but their fixed gating cannot implement the content-dependent operations required for transport or binding.
Together with Papers~I and~III, this yields a coherent trilogy instantiating the Structural Theorem on inference primitives and architectural realizability: Paper~I establishes \emph{existence and separation} (Lemma~1); this paper establishes \emph{mechanism} (Lemma~2)---showing that gradient dynamics reliably construct the primitives an architecture can support; Paper~III establishes \emph{scaling and completeness} (Lemma~3)---showing these mechanisms persist in production LLMs.
\bibliographystyle{ACM-Reference-Format}
\bibliography{references}
\end{document}